\documentclass[10pt,twocolumn,letterpaper]{article}

\PassOptionsToPackage{pdftex,unicode}{hyperref}

\def\CHK#1 {\textcolor{magenta}{{\bf [CHK:}~#1{\bf ]}}~}
\def\ADD#1 {\textcolor{cyan}{{\bf [To add:}~#1{\bf}]}~}

\newcommand{\ie}{\emph{i.e., }}
\newcommand{\eg}{\emph{e.g., }}

\usepackage[pagenumbers]{cvpr} 

\usepackage{graphicx}
\usepackage{amsmath}
\usepackage[accsupp]{axessibility}
\usepackage{amssymb}
\usepackage{booktabs}
\usepackage[noend]{algpseudocode}

\usepackage{algorithmicx,algorithm}

\usepackage{times}
\usepackage{pifont}       
\usepackage{bbding}       
\usepackage{setspace}
\usepackage{multirow}
\usepackage{cite}
\usepackage{comment}
\usepackage{booktabs}
\usepackage{arydshln}
\usepackage{color}
\usepackage{xcolor}
\usepackage{bm}
\usepackage[normalem]{ulem}
\usepackage{multirow}
\begin{document}


\title{IOTA: Corrective Knowledge-Guided Prompt Learning via Black-White Box Framework}

\author{
Shaokun Wang,
Yifan Yu,
Yuhang He,
Weili Guan,
Yihong Gong\\ 
Harbin Institute of Technology (Shenzhen), Xi'an Jiaotong University\\
{\tt\small wangshaokun@hit.edu.cn}, {\tt\small yyf1999@stu.xjtu.edu.cn} \\
 {\tt\small heyuhang@xjtu.edu.cn}, {\tt\small honeyguan@gmail.com}, {\tt\small ygong@mail.xjtu.edu.cn}}

\maketitle
\begin{abstract}
Recently, 
adapting pre-trained models to downstream tasks has attracted increasing interest. 
Previous Parameter-Efficient-Tuning (PET) methods regard the pre-trained model as an opaque Black Box model, relying purely on data-driven optimization and underutilizing their inherent prior knowledge. 
This oversight limits the models’ potential for effective downstream task adaptation.  
To address these issues, 
we propose a novel black-whIte bOx prompT leArning framework (IOTA), which integrates a data-driven Black Box module with a knowledge-driven White Box module for downstream task adaptation. 
Specifically, the White Box module derives corrective knowledge by contrasting the wrong predictions with the right cognition. This knowledge is verbalized into interpretable human prompts and leveraged through a corrective knowledge-guided prompt selection strategy to guide the Black Box module toward more accurate predictions. 
By jointly leveraging knowledge- and data-driven learning signals, IOTA achieves effective downstream task adaptation. 
Experimental results on 12 image classification benchmarks under few-shot and easy-to-hard adaptation settings demonstrate the effectiveness of corrective knowledge and the superiority of our method over state-of-the-art methods. 
\end{abstract}

\begin{figure}[t]
\centering
  
  	\includegraphics[width=1.0\linewidth]{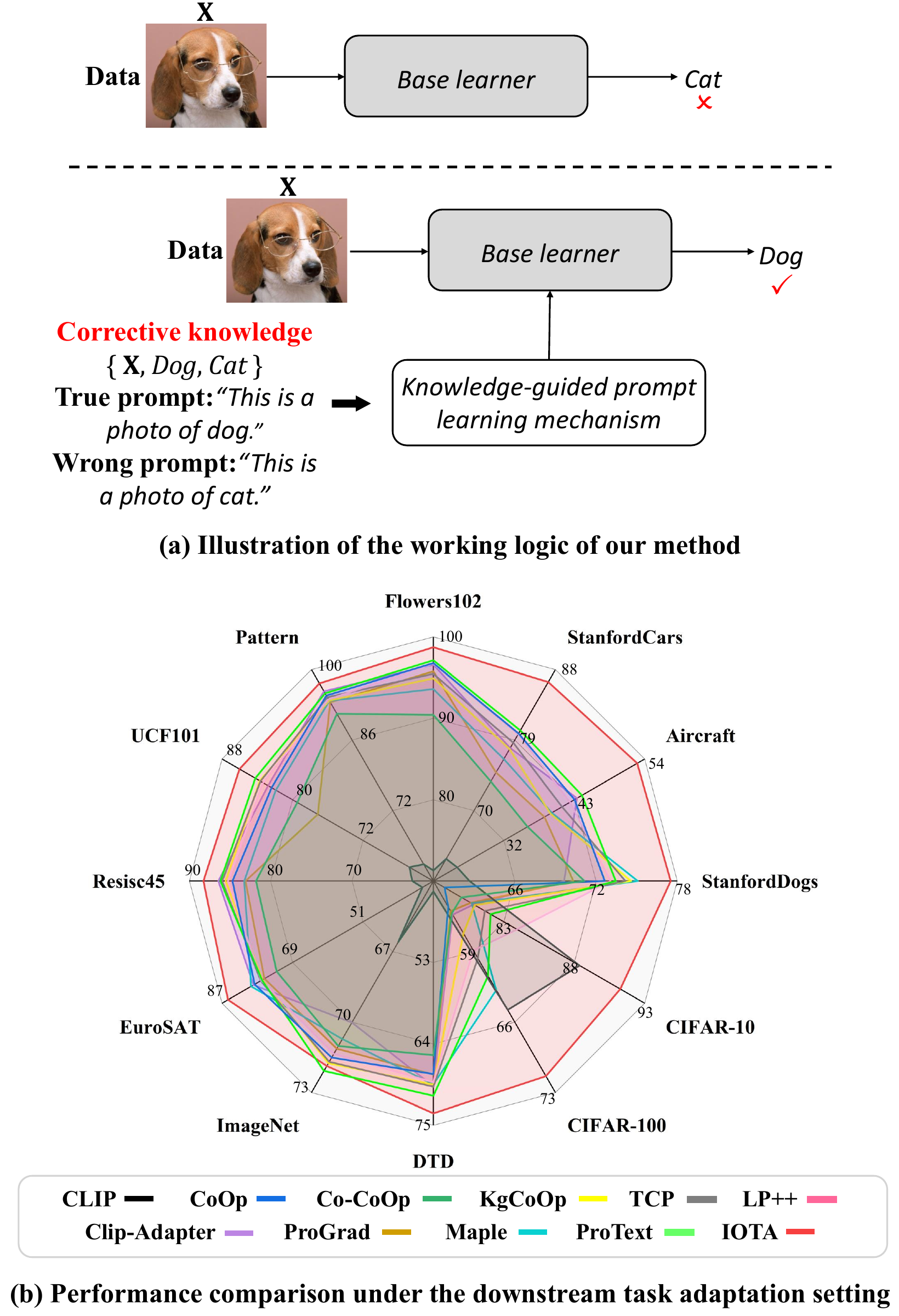}
   
  \caption
    {Comparison of existing Black Box methods and our Black-White Box method. (a) Our IOTA uses the knowledge-guided prompt learning mechanism (\ie White Box module) to correct the wrong prediction of the base learner (\ie Black Box module).  (b) Our IOTA surpasses the SOTA methods on 12 datasets. 
  }
  \label{fig:concept}
\end{figure}

\section{Introduction}
Existing large models~\cite{align, clip, gpt3} have achieved remarkable performance across a wide range of downstream tasks. 
The prevalent paradigm involves pre-training models on task-agnostic objectives and subsequently adapting them to specific downstream tasks.
However, full fine-tuning often disrupts the pre-trained latent space, leading to catastrophic forgetting of upstream knowledge~\cite{tl_survey2}, and requires storing separate parameter copies for each task. 

Parameter-Efficient-Tuning (PET) methods mitigate these issues by updating only a small subset of parameters. 
Representative approaches include prompt learning (\eg CoOp~\cite{coop}, Co-CoOp~\cite{cocoop}, and Maple~\cite{maple}) and adapter-based methods (\eg Clip-Adapter~\cite{clip-adapter}), which introduce task-specific parameters while keeping the pre-trained backbone frozen. 
Although effective, freezing the backbone treats the model as a Black Box by limiting access to its internal representations. This restriction reduces opportunities for knowledge-based analysis or correction. 
More specifically, without interpretable internal signals, it is difficult to diagnose errors, design corrective interventions, or directly leverage the model’s embedded priors~\cite{greybox1, bb_nature, greybox3}. 
Consequently, PET methods rely primarily on data-driven optimization and may not fully exploit the pre-trained model’s knowledge. 
By contrast, White Box models~\cite{greybox2,greybox6} provide explicit reasoning and interpretable representations through structured rules or symbolic components. 
However, such approaches often lack scalability and struggle to model high-dimensional data. 
This trade-off between interpretability and performance has motivated recent efforts~\cite{greybox4,greybox5} to explore hybrid Black-White Box frameworks that combine the complementary strengths of both paradigms.

Inspired by this insight,  we propose a black-whIte bOx prompT leArning framework (IOTA) to adapt pre-trained models to downstream tasks. 
IOTA consists of (\romannumeral1) \textbf{A data-driven Black Box module}: a pre-trained visual encoder serving as the base learner, which encodes rich prior knowledge acquired from upstream data; and
(\romannumeral2) \textbf{A knowledge-driven White Box module}: a knowledge-guided prompt learning mechanism that introduces human-understandable corrections to the base learner’s predictions. 
Specifically, we construct corrective knowledge according to the right cognition and the base learner's wrong predictions. 
These wrong predictions reveal aspects of the base learner’s prior knowledge that can be interpreted by humans. 
For example, 
as illustrated in Fig.~\ref{fig:concept} (a), consider a “dog” image $\mathbf{X}$ of the downstream task, where the base learner incorrectly predicts it as a “cat”. 
The corresponding corrective knowledge is defined as $\{\mathbf{X}, Dog, Cat\}$ and then verbalized into true-wrong prompts. 
These prompts form a knowledge prompt set, where each human prompt guides a learnable prompt to facilitate model correction. 
Afterwards,
a corrective knowledge-guided prompt selection strategy identifies effective prompts, which are fed as additional inputs to the base learner. 
By integrating the data-driven learning capacity of the Black Box and the knowledge-driven correction of the White Box, IOTA achieves more effective adaptation to downstream tasks.

We conduct comprehensive experiments on 12 image classification datasets, including 4 natural datasets, 4 fine-grained datasets, and 4 specialized datasets. 
In few-shot downstream task adaptation setting, as shown in Fig.~\ref{fig:concept} (b), IOTA significantly outperforms existing SOTA methods.  
Notably, in the more challenging easy-to-hard downstream task adaptation setting, IOTA improves significantly when facing hard samples with richer corrective knowledge.

In summary, 
the contributions in this work are listed as follows:
\begin{itemize}
    
    \item 
     We propose a novel Black-White Box prompt learning framework for the downstream task adaptation, integrating a data-driven Black Box module with a knowledge-driven White Box module. 
    
    \item We introduce corrective knowledge by contrasting right cognition with the base learner’s wrong predictions. This knowledge is verbalized into interpretable true-wrong prompts and applied via a corrective knowledge-guided prompt selection strategy. An easy-to-hard downstream task adaptation setting is designed to validate its effectiveness. 

    \item   
    Extensive experiments on 12 benchmark datasets show that IOTA consistently outperforms state-of-the-art methods in both few-shot and easy-to-hard downstream task adaptation settings. 
\end{itemize}

\section{Related Work}

\subsection{Visual Prompt Learning for Pre-trained Models}
Drawing inspiration from prompt learning techniques in the NLP field~\cite{p-tuning, power, prefix, zhong2021}, visual prompt learning has been introduced to CV to adapt pre-trained models to downstream tasks by fine-tuning a small number of task-specific parameters while keeping the backbone frozen~\cite{vpt,nlp_survey}. 
Similar to strong competitors such as adapter tuning~\cite{clip-adapter, tip-adapter, R-adapter}, and bias tuning~\cite{bias-t}, visual prompt learning offers comprehensive advantages in terms of both performance and per-task storage efficiency. 
Early works focused on vision-only models~\cite{vpt,l2p,nps}, with VPT~\cite{vpt} pioneering the exploration of visual prompt learning. 
Furthermore, 
several notable works have extended visual prompt learning to Vision-Language Models (VLMs)~\cite{evp, prograd, maple, cocoop, coop, pdl, ju2022, DPT, sgva_clip, TCP, lp24, ProText}. 
CLIP~\cite{clip} is a representative work that uses human prompts (\ie discrete prompts, like “a photo of a cat.”) to align visual and semantic modality, achieving impressive performance on zero-shot downstream task adaptation. 
However,
fixed human prompts are highly sensitive to results and are limited to flexibility and adjustability~\cite{pl_review}. 
Recent works such as CoOp~\cite{coop}, Co-CoOp~\cite{cocoop}, ProGrad~\cite{prograd}, Maple~\cite{maple}, TCP~\cite{TCP}, and  ProText~\cite{ProText} enhance the flexibility and robustness of visual prompt learning through learnable prompts. 
KgCoOp~\cite{kgcoop} further aligns human and learnable prompts via a knowledge-guided constraint to preserve generalization across tasks. 
However, learnable prompts remain uninterpretable, limiting understanding of how they influence model behavior and restricting adaptation to purely data-driven optimization.
In contrast, our method introduces corrective knowledge that bridges human-understandable cognition and model learning.
By verbalizing this knowledge as true–wrong human prompts, we guide the learning of corresponding learnable prompts.
A corrective knowledge-guided prompt selection strategy is then applied to choose suitable prompts for model correction.
Through this process, our method achieves explicitly correcting the model’s wrong predictions rather than merely aligning prompt types.

\subsection{Black-White Box Model Theory}
DNNs have achieved significant success in various fields. 
However, 
they are still regarded as unreliable Black Box models because their inner workings are unknown and difficult to interpret for humans~\cite{greybox2, greybox4}. 
Interpretability is critical in many real-world applications, where clear explanations of the modeling process, intermediate results, and final outputs are required for users without technical expertise~\cite{tcsvt3,tcsvt4,tcsvt5,tcsvt6}.  

In contrast to Black Box models, knowledge-driven White Box models have transparent inner workings, programming steps, and prediction-making mechanisms~\cite{greybox2,tkde5}. 
This transparency enables White Box models to provide interpretable explanations to humans. 
Algorithms such as decision trees~\cite{DT} and rule-based systems~\cite{RBS} exemplify this approach. 
Nevertheless, 
due to the difficulty in acquiring expert knowledge and establishing comprehensive rule-based systems, White Box models often struggle with scalability and accuracy in complex scenarios~\cite{greybox4}. 

To leverage the complementary strengths of both paradigms, recent studies have explored hybrid Black–White Box models~\cite{greybox4,greybox5}.  
The Black-White Box model has the obvious advantage over the White Box model in terms of accuracy, since it has a Black Box model as its base predictor, while the interpretation of the predictions is achieved by designing a knowledge-driven White Box model~\cite{greybox1}. 
For example, 
SHAP~\cite{shap} explains the predictions of the Black Box model by evaluating the contribution of each feature to the model’s output using the knowledge of game theory and Shapley values. 
It effectively establishes the connection between knowledge and the Black Box model, addressing a key challenge that many current studies~\cite{tcsvt1,tcsvt2,tcsvt7,tcsvt8,tcsvt9} are working to solve. 
Motivated by these insights, our method incorporates corrective knowledge into the Black Box module via a knowledge-guided prompt learning mechanism, serving as the White Box module. This integration leverages human-understandable knowledge to enhance the model’s adaptation to downstream tasks.

\begin{figure*}[ht]
	\centering
	\includegraphics[width=0.9\textwidth]{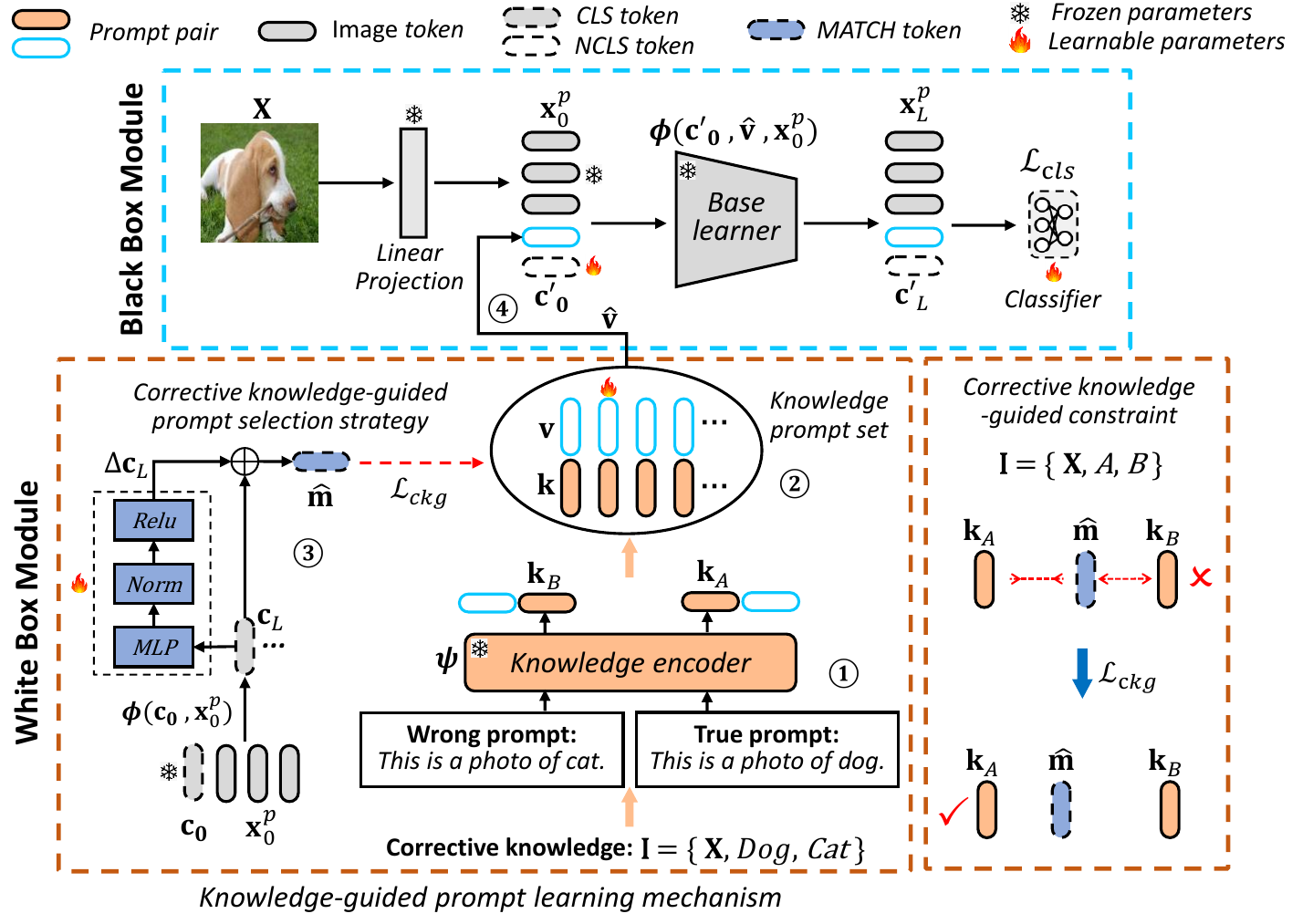} 
	\caption{
	An overview of our IOTA framework.
The corrective knowledge is leveraged to design a knowledge-guided prompt learning mechanism (\ie the White Box module), which provides interpretable guidance to enhance the performance of the base learner (\ie the Black Box module). 
	}
	\label{figs:framework}
\end{figure*}

\section{Methodology}
\subsection{Overall} 
In our framework, a \textit{pre-trained visual encoder} is used as the \textit{base learner} $\phi$, and a \textit{pre-trained semantic encoder} is used as the \textit{knowledge encoder} $\psi$.
As shown in Fig.~\ref{figs:framework}, IOTA includes two parts: a data-driven \textit{Black Box module} and a knowledge-driven \textit{White Box module}.
The \textit{base learner} $\phi$ acts as the \textit{Black Box module}.
The \textit{White Box module} is a \textit{knowledge-guided prompt learning mechanism} that contains a \textit{knowledge encoder}, a \textit{knowledge prompt set}, and a \textit{corrective knowledge-guided prompt selection strategy}.

\subsection{Black Box Module: Base Learner}
Fine-tuning pre-trained encoders on downstream tasks often disturbs the learned feature space and causes forgetting of upstream knowledge. 
To prevent this, we freeze the pre-trained visual encoder and designate it as the Black Box module, which provides stable yet opaque representations for downstream adaptation. 
This design ensures that the subsequent White Box module can focus on correcting decision-level errors rather than relearning visual representations.
Specifically, the base learner $\phi$ is a Vision Transformer (ViT) with $L$ transformer layers. 
Given an image $\mathbf{X}$, it is divided into $E$ patches and embedded into patch tokens $\mathbf{x}^{p} \in \mathbb{R}^{E \times d_t}$. 
These tokens, together with a learnable CLS token $\mathbf{c}_{0}$, are processed by the transformer blocks:
\begin{equation}
    [\mathbf{c}_{i}, \mathbf{x}^{p}_{i}] = \phi_{i}([\mathbf{c}_{i-1}, \mathbf{x}^{p}_{i-1}]), \quad i=1,\dots,L.
\end{equation}
The final CLS token $\mathbf{c}_{L}$ serves as the image representation for downstream classification. 
These frozen representations are subsequently refined by the knowledge-guided prompts in the White Box module, enabling effective correction of the model’s wrong predictions. 

\subsection{White Box Module: Knowledge-Guided Prompt Learning Mechanism}
Built upon the frozen Black Box module, we introduce a knowledge-driven White Box module that explicitly incorporates human-understandable corrective knowledge to guide downstream adaptation. 
This module employs a \textit{knowledge-guided prompt learning mechanism} to bridge right cognition and model predictions. 
Concretely, we first formulate \textit{corrective knowledge triplets} and verbalize them into interpretable true-wrong prompts. 
These prompts are integrated into a task-specific \textit{knowledge prompt set}, serving as a bridge between semantic understanding and model behavior. 
A \textit{corrective knowledge-guided prompt selection strategy} is then developed to dynamically select suitable prompts for each input sample. 
The selected prompts act as interpretable knowledge carriers that transfer the right cognition to the base learner, thereby correcting its wrong predictions and achieving knowledge-guided adaptation.
\subsubsection{Formulation and Verbalization of Corrective Knowledge} 
For each sample $\mathbf{X}$ in a downstream task, we obtain the prediction from the base learner and the corresponding ground-truth label, which represents the right cognition. 
We formalize the corrective knowledge as a triplet:
\begin{equation}
    \mathbf I=\{\mathbf{X},A,B\},
    \label{eq_I}
\end{equation}
where $A$ denotes the true class index and $B$ denotes the predicted class index. 
When $A = B$, the prediction is correct; otherwise, $A \neq B$ indicates a wrong prediction, reflecting the wrong cognition of the base learner. 
Here, corrective knowledge refers to human-interpretable signals derived from the discrepancy between predictions and ground-truth labels. 
Unlike raw training samples, these signals are symbolic and reusable in all instances, providing semantic guidance to correct decision-level errors rather than serving as direct supervision.  
Inspired by human cognitive theories suggesting that people refine their decision-making by contrasting correct and incorrect judgments~\cite{human}, 
our framework aims to enhance right cognition while suppressing wrong cognition.

To enable human-understandable guidance, we verbalize each corrective knowledge triplet $\mathbf I=\{\mathbf{X},A,B\}$ into interpretable prompts. 
When $A \neq B$, we define the corresponding true-wrong prompts $\mathbf{Z}^{T/W}$ as:
\begin{equation}
\begin{cases}
\mathbf{Z}^{T} = \textit{This is a photo of a [$CLASS_{A}$]}. \\
\mathbf{Z}^{W} = \textit{This is a photo of a [$CLASS_{B}$]}.
\end{cases}
\end{equation}
The true prompt expresses correct semantic knowledge, whereas the wrong prompt illustrates the model’s misperceived prior knowledge in an interpretable form. 
For samples that are correctly classified ($A = B$), only the true prompt is used. 
Note that prompt templates can be flexibly adjusted across different downstream tasks.

\subsubsection{Knowledge Prompt Set Construction} 
For each downstream task, we integrate all true and wrong human prompts into a unified knowledge prompt set $\mathcal{V}$.
Each human prompt $\mathbf{Z}_j$ is first tokenized and mapped to word embeddings by the knowledge encoder $\psi$, which then processes them through a Transformer.
The representation at the last token position is layer-normalized and projected to obtain the knowledge embedding $\mathbf{k}_j$:
\begin{equation}
\mathbf{k}_j = \psi(\mathbf{Z}_j).
\end{equation}
The knowledge prompt set is then defined as
$\mathcal{V}=\{(\mathbf k_{1},\mathbf v_{1}),(\mathbf k_{2},\mathbf v_{2}),...,(\mathbf k_{S},\mathbf v_{S})\}$,
where $S$ is the number of prompt pairs in the set. 
Each pair $(\mathbf{k}_j, \mathbf{v}_j)$ consists of a fixed knowledge embedding $\mathbf{k}_j$ and a learnable prompt $\mathbf{v}_j$, which are aligned in a one-to-one manner.
Thus, any true-wrong prompt pair $<\mathbf{Z}^T, \mathbf{Z}^W>$ can be represented as the embedding pair $<\mathbf{k}_A, \mathbf{k}_B>$ within $\mathcal{V}$.

\subsubsection{Corrective Knowledge-Guided Prompt Selection Strategy} 
\paragraph{MATCH Token Design}
To bridge the distributional gap between pre-trained representations and downstream tasks, we introduce a lightweight \textit{MATCH token}:
\begin{equation}
\hat{\mathbf{m}} = \mathbf{c}_{L} + \Delta \mathbf{c}_{L}, \quad
\Delta \mathbf{c}_{L} = \mathrm{RELU}\big(\mathrm{NORM}(\mathrm{MLP}(\mathbf{c}_{L}))\big).
\end{equation}
This design allows $\hat{\mathbf{m}}$ to better align with the downstream feature distribution while preserving information from the upstream CLS token $\mathbf{c}_{L}$.

\paragraph{Soft Prompt Matching} 
Instead of selecting a single closest prompt, we compute a softmax-weighted similarity over the entire knowledge prompt set $\mathcal{V}$:
\begin{equation}
P(\mathbf{k}_j | \hat{\mathbf{m}}) = \frac{\exp(Q(\hat{\mathbf{m}}, \mathbf{k}_j)/\tau)}{\sum_{i=1}^{S} \exp(Q(\hat{\mathbf{m}}, \mathbf{k}_i)/\tau)},
\end{equation}
where $Q(\hat{\mathbf{m}}, \mathbf{k}_j)$ is the cosine similarity between $\hat{\mathbf{m}}$ and $\mathbf{k}_j$, $S$ is the number of prompt pairs in the set, and $\tau$ is a temperature hyperparameter. 
This formulation defines a probability distribution over all prompts, encouraging the MATCH token to learn a smooth semantic mapping across the prompt space.

\paragraph{Corrective Knowledge-Guided Constraint} 
To incorporate corrective knowledge, we define the training loss as:
\begin{equation}
\mathcal{L}_{ckg} = -\log P(\mathbf{k}_A \mid \hat{\mathbf{m}}) - \mathbb{I}(A \neq B) \cdot \log \big( 1 - P(\mathbf{k}_B \mid \hat{\mathbf{m}}) + \epsilon \big),
\label{loss}
\end{equation}
where $\mathbf{k}_A$ and $\mathbf{k}_B$ correspond to the true and wrong prompts, respectively, $\mathbb{I}(\cdot)$ is the indicator function, which returns 1 if its argument is true and 0 otherwise, and $\epsilon$ is a small constant ($1e{-8}$) to ensure numerical stability. 
This constraint encourages the MATCH token to assign high probability to the true prompt while reducing the probability of the wrong prompt, effectively transferring corrective knowledge into the prompt selection process.

\paragraph{Prompt Selection} 
We adopt a soft-matching strategy to compute a weighted combination of all prompts in the knowledge prompt set $\mathcal{V}$:
\begin{equation}
\hat{\mathbf{v}} = \sum_{(\mathbf{k}_j, \mathbf{v}_j) \in \mathcal{V}} w_j \cdot \mathbf{v}_j, \quad
w_j = \frac{\exp(Q(\hat{\mathbf{m}}, \mathbf{k}_j)/\tau)}{\sum_{(\mathbf{k}_{j'},\mathbf{v}_{j'}) \in \mathcal{V}} \exp(Q(\hat{\mathbf{m}}, \mathbf{k}_{j'})/\tau)},
\end{equation}
where $Q(\hat{\mathbf{m}}, \mathbf{k}_j)$ is the cosine similarity between the MATCH token $\hat{\mathbf{m}}$ and the knowledge embedding $\mathbf{k}_j$, and $\tau$ is a temperature parameter controlling the smoothness of the softmax. 
To prevent $\hat{\mathbf{v}}$ from exhibiting excessively large magnitude and ensures numerical stability, we apply $\ell_2$ normalization to each $\mathbf{v}_j$ before aggregation. 
By softly weighting multiple candidate prompts, the model captures inter-prompt relationships and reduces the training-inference discrepancy. 
Even without access to the true prompt during inference, $\hat{\mathbf{m}}$ guides the base learner toward semantically appropriate prompts, ensuring effective corrective guidance for downstream task adaptation. 
\subsection{Optimization for the Black-White Box Framework}

The prompt embedding $\hat{\mathbf{v}}$ is concatenated with image tokens and fed into our Black Box module. 
We introduce a trainable NCLS token $\mathbf{c'}$, initialized from the original CLS token $\mathbf{c}$ but trained independently for downstream task adaptation.  
Specifically, for the $i$-th transformer block $\phi_i$:

\begin{equation}
[\mathbf{c'}_i, \hat{\mathbf{v}}_i, \mathbf{x}^p_i] = \phi_i([\mathbf{c'}_{i-1}, \hat{\mathbf{v}}_{i-1}, \mathbf{x}^p_{i-1}]), \quad i = 1,2,\dots,L,
\end{equation}
where $\hat{\mathbf{v}}_i$ denotes the injected corrective prompt embedding. 
After the final layer, the NCLS token $\mathbf{c'}_L$ is fed into the classifier $h(\cdot)$ to predict the downstream label $y$.

We optimize the Black-White Box framework end-to-end using a multi-objective loss:

\begin{equation}
\mathcal{L} = \mathcal{L}_{cls}(h(\mathbf{c'}_L), y) + \lambda \mathcal{L}_{ckg},
\end{equation}
where $\mathcal{L}_{cls}$ is the cross-entropy loss for classification, and $\mathcal{L}_{ckg}$ is the corrective knowledge-guided loss that trains the prompt selection mechanism (see Eq.~\ref{loss}). 
The hyper-parameter $\lambda$ balances the contributions of classification and knowledge-guided prompt selection.

\section{Learning Knowledge from Easy to Hard}
To demonstrate the effectiveness of corrective knowledge, we propose an easy-to-hard downstream task adaptation setting, which consists of two stages: an easy curriculum and a hard curriculum.

In the easy curriculum, we select easy samples that are representative of their classes to provide a stable starting point for adapting the base learner. In the hard curriculum, we focus on hard samples, which are more likely to be wrong predicted and thus contain richer corrective knowledge to further enhance the base learner. 

We quantify sample difficulty based on its distance to the class centroid in the feature space of the base learner. Specifically, for a sample $\mathbf{X}$ of class $j$, the difficulty score $\mathbf{D}$ is defined as:
\begin{equation}
\mathbf{D} = dis(\mathbf{e}^{x}, \mathbf{o}_{j}),
\end{equation}
where $\mathbf{e}^{x}$ is the feature of $\mathbf{X}$, $\mathbf{o}_{j}$ is the centroid of class $j$, and $dis(\cdot, \cdot)$ denotes cosine distance. A smaller distance indicates an easier (more representative) sample, while a larger distance corresponds to a harder sample that is more likely to be wrong classified.

During training, we first select the $N$ easiest samples per class to train the IOTA-e model. Next, using IOTA-e as the updated base learner, we select the $N$ hardest samples per class to train the IOTA-h model. Both IOTA-e and IOTA-h share the same framework. In each stage, only $N$ samples per class are used (\eg 4, 8, or 16).

If IOTA-h achieves significantly better performance than IOTA-e, it validates that our knowledge-guided prompt learning mechanism can leverage richer corrective knowledge from hard samples to further improve the base learner.

\begin{table*}[]
\caption{Datasets statistics. }
\centering
\scalebox{0.88}{
\begin{tabular}{@{}lcrrrc@{}}
\toprule[1.5pt]
Dataset        & Description                   & Classes & Train   & Test  & Human prompt template \\ \midrule
Flowers102 & \multirow{4}{*}{Fine-grained} & 102  & 5726    & 2463  &  “This is a photo of a [CLASS], a type of flower.” \\
Stanford Cars  &  & 196 & 8144    & 8041  &  “This is a photo of a [CLASS].” \\
Aircraft  &   & 100 & 6667  & 3333  &  “This is a photo of a [CLASS], a type of aircraft.” \\
Stanford Dogs  &  & 120  & 12000 & 8580 & “This is a photo of a [CLASS], a type of dog.”  \\ \midrule 
CIFAR-10  & \multirow{4}{*}{Natural}  & 10  & 50000   & 10000 &   “This is a photo of a [CLASS].”  \\
CIFAR-100  &   & 100     & 50000   & 10000 &  “This is a photo of a [CLASS].”  \\
DTD  &   & 47  & 3760    & 1880  & “This is a photo of a [CLASS].” \\
ImageNet &   & 1000    & 1281166 & 50000 & “This is a photo of a [CLASS].”   \\ \midrule
EuroSAT  & \multirow{4}{*}{Specialized}  & 10      & 18900   & 8100  &   “This is a centered satellite photo of [CLASS].” \\
Resisc45  &  & 45  & 6300  & 25200 & “This is an aerial imagery of a [CLASS].”  \\
Pattern  &  & 38      & 24320   & 6080  &  “This is an aerial imagery of a [CLASS].” \\
UCF  &   & 101 & 9537    & 3783  & “This is a photo of a person doing [CLASS].”  \\ \bottomrule[1.5pt]
\end{tabular}}
\label{tbl:a1}
\end{table*}

\begin{table*}[tbp]
\caption{  Comparison with the SOTA methods on 12 datasets under the \textbf{few-shot downstream task adaptation setting}, where the base learner is \textbf{ViT-B/16}. Our IOTA has achieved the highest average accuracy. The highest accuracy on each dataset is indicated in bold. }
\centering
\scalebox{0.75}{
\begin{tabular}{l|ccccccccccccc|c}
\toprule[1.5pt]
 & \multicolumn{13}{c|}{\textbf{Dataset}} &  \\ \cline{2-14} 
\multirow{5}{*}{\textbf{Method}} & \multicolumn{1}{c}{\multirow{5}{*}{\rotatebox{60}{Flowers102}}} & \multirow{5}{*}{\rotatebox{60}{StanfordCars}} & \multirow{5}{*}{\rotatebox{60}{Aircraft}} & \multirow{5}{*}{\rotatebox{60}{StanfordDogs}} & \multirow{5}{*}{\rotatebox{60}{CIFAR-10}} & \multirow{5}{*}{\rotatebox{60}{CIFAR-100}} & \multirow{5}{*}{\rotatebox{60}{DTD}} & \multirow{5}{*}{\rotatebox{60}{ImageNet}} & \multirow{5}{*}{\rotatebox{60}{EuroSAT}} & \multirow{5}{*}{\rotatebox{60}{Resisc45}} & \multirow{5}{*}{\rotatebox{60}{UCF101}} & \multicolumn{1}{l|}{\multirow{5}{*}{\rotatebox{60}{Pattern}}} & \multirow{5}{*}{\rotatebox{60}{Average}} &\multirow{5}{*}{\textbf{Shot}} \\
& \multicolumn{1}{c}{} &  &  &  &  &  &  & &  &  &   & \multicolumn{1}{l|}{}  &  \\
& \multicolumn{1}{c}{}  &  & &   &   &   &   &   &   &    &    & \multicolumn{1}{l|}{} &  \\
& \multicolumn{1}{c}{} &  &  &  &  &  &  & &   &  &    & \multicolumn{1}{l|}{} &   \\
& \multicolumn{1}{c}{}  &  &   &   &   &  &  &  &   &  &  & \multicolumn{1}{l|}{}  &  \\ \hline
CLIP  & 71.30 & 63.84 & 24.72 & 62.67 & 88.38 & 64.78 & 43.40  & 66.59 & 35.80  &  62.60  & 66.72  & \multicolumn{1}{l|}{61.33} & 59.34 & 0 \\
\hline

Co-CoOp  & 90.42 & 73.72 & 35.70 & 71.15 & 80.00 & 55.19 & 65.53 & 71.02 & 73.02  & 81.81 & 79.46 & \multicolumn{1}{l|}{91.23} &  72.35 &\multirow{10}{*}{\textbf{16-shot}}\\

ProGrad  & 95.78 &  74.87 &  38.34  & 70.33  &  80.66 & 54.97 &  68.14 &71.13 & 76.16 & 83.11  & 77.13  & \multicolumn{1}{l|}{93.52} &  73.68    \\

Clip-Adapter & 96.59  & 77.99 &  43.41  & 69.64 & 80.86  & 55.33 & 69.73  &  70.02 &  78.96   & 86.44 & 82.77 & \multicolumn{1}{l|}{95.71} & 75.62 \\

CoOp  & 96.79  &  79.84  & 43.05 & 72.65 & 78.80 & 54.57 &  68.09  &  71.51 &  78.68 &  84.70 &  82.37  & \multicolumn{1}{l|}{94.85} &  75.49  \\

Maple  & 93.59  & 76.58 & 39.48 & 75.08  & 80.72  & 62.84 & 69.47 &  70.72  & 79.57 &  83.26  &  81.89  & \multicolumn{1}{l|}{93.82} & 75.59   \\ 
KgCoOp  & 94.92  & 77.85 & 39.27 & 74.59  & 80.94  & 57.18 & 69.52 &  71.75  & 76.68 &  85.54  &  83.72  & \multicolumn{1}{l|}{93.87} & 75.48   \\ 

TCP  & 95.46  & 78.66 & 41.09 & 74.17  & 81.65  & 59.67 & 69.80 &  71.69  & 77.13 &  85.97  &  83.64  & \multicolumn{1}{l|}{94.53} & 76.12   \\

LP++  & 96.71  & 79.75 & 42.33 & 73.08  & 83.93  & 58.90 & 69.19 &  71.35  & 76.99 &  85.26  &  83.35  & \multicolumn{1}{l|}{94.11} & 76.25   \\

ProText  & 97.12  & 80.23 & 44.21 & 73.46  & 82.06  & 61.42 & 71.01 &  \textbf{72.08}  & 76.80 &  86.18  &  84.22  & \multicolumn{1}{l|}{95.30} & 77.01   \\
\cline{1-14}

IOTA & \textbf{98.74} & \textbf{86.41} & \textbf{52.84} & \textbf{77.51} & \textbf{91.27} & \textbf{71.37} & \textbf{73.40} & 71.87 & \textbf{85.49} & \textbf{88.27} & \textbf{86.02} &  \multicolumn{1}{l|}{ \textbf{97.27}}  & \textbf{81.71} \\
\hline

Co-CoOp  & 88.02 & 71.51 & 32.46 & 70.09 & 78.50 & 55.03 & 59.15 & 70.19 & 65.20  & 78.06 & 76.79 & \multicolumn{1}{l|}{86.91} &  69.33 &\multirow{10}{*}{\textbf{8-shot}}\\

ProGrad  & 93.54 &  73.37 &  34.53  & 69.78  &  79.08 & 53.50 &  62.93 & 70.44& 69.74 & 80.29  & 76.08  & \multicolumn{1}{l|}{90.84} &   71.18   \\

Clip-Adapter & 94.03  & 73.64 &  36.57  & 66.60 & 79.35  & 53.95 & 64.15  & 69.02  &  71.99   &  82.29 & 80.70 & \multicolumn{1}{l|}{92.50} & 72.07 \\

CoOp  & 95.01  & 76.50  & 36.57 & 70.01 & 78.67 & 52.97 &  64.52  &  69.19 &  70.16 &  81.38 &  79.88  & \multicolumn{1}{l|}{91.45} &  72.19  \\

Maple  & 90.05  & 72.62 & 32.91 & \textbf{72.45}  & 79.99 &  61.00 & 64.89 &  70.06 & 66.98 &  79.07  &  79.12                & \multicolumn{1}{l|}{86.97} &    71.34 \\ 

KgCoOp  & 94.28  & 76.46 & 36.69 & 71.43  & 80.08  & 55.38 & 65.96 &  \textbf{70.68}  & 72.81 &  81.94  &  81.36  & \multicolumn{1}{l|}{91.02} & 73.17   \\ 

TCP  & 94.47  & 76.30 & 36.54 & 71.61  & 79.82  & 56.49 & 66.50 &  70.56  & 72.76 &  82.40  &  81.53  & \multicolumn{1}{l|}{91.61} & 73.38   \\ 

LP++  & 93.96  & 75.39 & 36.99 & 71.05  & 80.26  & 58.73 & 66.84 &  69.57  & 73.69 &  80.59  &  80.92  & \multicolumn{1}{l|}{91.44} & 73.29   \\ 

ProText  & 95.45  & 75.73 & 38.52 & 71.77  & 80.63  & 59.36 & 67.27 &  70.40  & 72.22 &  82.31  &  81.08  & \multicolumn{1}{l|}{93.25} & 74.00   \\ 
\cline{1-14}

IOTA & \textbf{97.48} & \textbf{80.82} & \textbf{40.53} & 69.90 & \textbf{83.22} & \textbf{63.07} & \textbf{69.73} & 69.95 & \textbf{76.72} & \textbf{84.05} & \textbf{81.89} &  \multicolumn{1}{l|}{ \textbf{96.17}}  & \textbf{76.13} \\ 
\bottomrule[1.5pt]
\end{tabular}}
\label{tbl:e1}
\end{table*}

\begin{table}[]
\caption{  Comparison with the SOTA methods on 3 kinds of datasets (4 fine-grained datasets, 4 natural datasets, and 4 specialized datasets) under the \textbf{few-shot downstream task adaptation setting}, where the base learner is \textbf{ViT-B/32}. Our IOTA has achieved the highest average accuracy. The highest accuracy on each kind of dataset is indicated in bold. }
\centering
\scalebox{0.75}{
\begin{tabular}{l|cccc|c}
\toprule[1.5pt]
 & \multicolumn{4}{c|}{\textbf{Dataset}} &  \\ \cline{2-5} 
\multirow{5}{*}{\textbf{Method}} & \multicolumn{1}{c}{\multirow{5}{*}{\rotatebox{60}{Fine-grained (4)}}} & \multirow{5}{*}{\rotatebox{60}{Natural (4)}} &\multicolumn{1}{l|}{\multirow{5}{*}{\rotatebox{60}{Specialized (4)}}}  & \multirow{5}{*}{\rotatebox{60}{Average}} &\multirow{5}{*}{\textbf{Shot}} \\
& \multicolumn{1}{c}{}   &   & \multicolumn{1}{l|}{}  &  \\
& \multicolumn{1}{c}{}  &  & \multicolumn{1}{l|}{} &  \\
& \multicolumn{1}{c}{} &  &     \multicolumn{1}{l|}{} &   \\
& \multicolumn{1}{c}{}  &  &  \multicolumn{1}{l|}{}  &  \\ \hline
CLIP  & 49.96 & 63.86 & \multicolumn{1}{l|}{51.90} & 55.24 & 0 \\
\hline

Co-CoOp  & 59.41 & 63.20  & \multicolumn{1}{l|}{78.65} & 67.09  &\multirow{10}{*}{\textbf{16-shot}}\\

ProGrad  & 63.91 &  64.48 & \multicolumn{1}{l|}{79.52} & 69.30  \\

Clip-Adapter & 65.85 & 63.77 & \multicolumn{1}{l|}{83.85} & 71.16 \\

CoOp  &65.95  & 64.68 & \multicolumn{1}{l|}{82.86} & 71.16 \\

KgCoOp  & 64.23 & 65.92 & \multicolumn{1}{l|}{82.11} &  70.75   \\ 

Maple  & 64.49 & 68.37 & \multicolumn{1}{l|}{83.32} &  72.06   \\ 

TCP  &65.14  & 67.73 & \multicolumn{1}{l|}{82.91} & 71.93 \\

LP++  & 64.65 & 66.63 & \multicolumn{1}{l|}{82.48} &  71.25   \\ 

ProText  & 67.39 & 68.06 & \multicolumn{1}{l|}{83.44} &  72.96   \\
\cline{1-5}

IOTA & \textbf{72.07} & \textbf{72.24}  &  \multicolumn{1}{l|}{ \textbf{85.87}}  & \textbf{76.73} \\
\hline

Co-CoOp  & 55.91 & 62.08  & \multicolumn{1}{l|}{74.23} &  64.08 &\multirow{10}{*}{\textbf{8-shot}}\\

ProGrad  & 60.97 & 62.59  & \multicolumn{1}{l|}{74.02} &  65.86 \\

Clip-Adapter & 60.91 & 62.72  & \multicolumn{1}{l|}{79.05} & 67.56  \\

CoOp  & 61.37 & 62.17  & \multicolumn{1}{l|}{78.19} & 67.24  \\

KgCoOp  & 60.15 & 64.02  & \multicolumn{1}{l|}{79.63} & 67.93  \\

Maple  & 60.01 & 65.47 & \multicolumn{1}{l|}{77.30} & 67.59  \\ 

TCP  &60.73  & 63.81 & \multicolumn{1}{l|}{78.97} & 67.84 \\

LP++  & 62.27 & 62.85 & \multicolumn{1}{l|}{79.70} &  68.27   \\ 

ProText  & 62.66 & 64.15 & \multicolumn{1}{l|}{79.46} &  68.76   \\
\cline{1-5}

IOTA & \textbf{64.75} & \textbf{67.17} &  \multicolumn{1}{l|}{ \textbf{82.82}}  & \textbf{71.58} \\ 
\bottomrule[1.5pt]
\end{tabular}}
\label{tbl:a4}
\end{table}

\begin{table*}[htbp]
\caption{ Comparison with the SOTA methods on 12 datasets under the \textbf{easy-to-hard downstream task adaptation setting}, where the base learner is \textbf{ViT-B/16}. Methods in the form of \textbf{‘xxx-e’} and \textbf{‘xxx-h’} represent \textbf{being trained only in the easy curriculum stage} and \textbf{being trained in the easy curriculum stage followed by the hard curriculum stage}, respectively. The highest accuracy in the easy/hard curriculum on each dataset is indicated with an underline / in bold. Absolute improvements from the easy curriculum to the hard curriculum are indicated in parentheses. }
\centering
\scalebox{0.75}{
\begin{tabular}{l|ccccccccccccc|c}
\toprule[1.5pt]
 & \multicolumn{13}{c|}{\textbf{Dataset}} &  \\ \cline{2-14} 
\multirow{5}{*}{\textbf{Method}} & \multicolumn{1}{c}{\multirow{5}{*}{\rotatebox{60}{Flowers102}}} & \multirow{5}{*}{\rotatebox{60}{StanfordCars}} & \multirow{5}{*}{\rotatebox{60}{Aircraft}} & \multirow{5}{*}{\rotatebox{60}{StanfordDogs}} & \multirow{5}{*}{\rotatebox{60}{CIFAR-10}} & \multirow{5}{*}{\rotatebox{60}{CIFAR-100}} & \multirow{5}{*}{\rotatebox{60}{DTD}} & \multirow{5}{*}{\rotatebox{60}{ImageNet}} & \multirow{5}{*}{\rotatebox{60}{EuroSAT}} & \multirow{5}{*}{\rotatebox{60}{Resisc45}} & \multirow{5}{*}{\rotatebox{60}{UCF101}} & \multicolumn{1}{l|}{\multirow{5}{*}{\rotatebox{60}{Pattern}}} & \multirow{5}{*}{\rotatebox{60}{Average}} &\multirow{5}{*}{\textbf{Shot}} \\
& \multicolumn{1}{c}{} &  &  &  &  &  &  & &  &  &   & \multicolumn{1}{l|}{}  &  \\
& \multicolumn{1}{c}{}  &  & &   &   &   &   &   &   &    &    & \multicolumn{1}{l|}{} &  \\
& \multicolumn{1}{c}{} &  &  &  &  &  &  & &   &  &    & \multicolumn{1}{l|}{} &   \\
& \multicolumn{1}{c}{}  &  &   &   &   &  &  &  &   &  &  & \multicolumn{1}{l|}{}  &  \\ \hline
CLIP  & 71.30 & 63.84 & 24.72 & 62.67 & 88.38 & 64.78 & 43.40  & 66.59 & 35.80  &  62.60  & 66.72  & \multicolumn{1}{l|}{61.33} & 59.34 & 0 \\
\hline
CoOp-e          & - & - & 44.34 & 75.59 & 82.50 & 58.72 & - &  71.30 & 80.84  & 86.29 & - & \multicolumn{1}{l|}{94.74} &  74.29 &\multirow{14}{*}{\textbf{16/stage}}\\
CoOp-h         & - & - & 45.33 & 75.19 & 78.80 & 58.02 & - &  70.75 & 64.69  & 85.35 & - & \multicolumn{1}{l|}{89.59} &  70.97 (-3.32) \\
Co-CoOp-e       & - & - & 35.43 & 71.68 & 82.54 & 56.29 & - &  71.27      & 72.32  & 82.05 & - & \multicolumn{1}{l|}{90.41} &  70.25 \\
Co-CoOp-h      & - & - & 37.02 & 71.95 & 81.52 & 54.95 & - & 71.01  & 60.72  & 82.30 & - & \multicolumn{1}{l|}{89.85} &  68.67 (-1.58) \\
Clip-Adapter-e  & - & - & 46.56 & 71.74 & 82.26 & 56.27 & - &  70.23 & 81.27  & \underline{87.70} & -  & \multicolumn{1}{l|}{95.86}  &  73.99 \\
Clip-Adapter-h & - & - & 47.37 & 71.67 & 82.23 & 56.55 & - & 70.00  & 81.26  & 87.10 & -  & \multicolumn{1}{l|}{96.05}  &  74.03 (+0.04) \\
ProGrad-e       & - & - & 39.48 & 71.00 & 80.26 & 55.99 & - &   71.68& 79.89  & 84.20 & -  & \multicolumn{1}{l|}{94.26} &   72.10   \\
ProGrad-h      & - & - & 43.74 & 74.95 & 79.88 & 57.86 &  -&  70.59 & 74.95  & 84.55 & -  & \multicolumn{1}{l|}{93.03} &  72.44 (+0.34)    \\
Maple-e         & - & - & 38.85 & 75.41 & 81.81 & 62.41 & - & 72.21  & 82.41  & 84.11 &  - & \multicolumn{1}{l|}{93.24} & 73.81    \\ 
Maple-h        & - & - & 41.31 & 75.52 & 80.79 & 62.36 & - & 71.24  & 85.16  & 84.31 &  - & \multicolumn{1}{l|}{94.08} &  74.35 (+0.54)   \\ 
KgCoOp-e  & - & - & 41.16 & 74.76 & 82.80 & 58.74 & - & 71.68  & 78.62  & 86.40 &  - & \multicolumn{1}{l|}{93.90} & 73.51    \\ 
KgCoOp-h  & - & - & 40.59 & 74.74 & 83.12 & 58.70 & - & 71.04  & 78.75  & 86.23 &  - & \multicolumn{1}{l|}{93.82
} & 73.37 (-0.14)   \\ 

TCP-e       & - & - & 41.82 & 75.86 & 82.36 & 59.88 & - &   71.78& 79.13  & 85.72 & -  & \multicolumn{1}{l|}{93.67} &   73.78   \\
TCP-h      & - & - & 42.37 & 75.24 & 82.91 & 60.31 &  -&  72.29 & 80.47  & 85.09 & -  & \multicolumn{1}{l|}{94.18} &  74.11 (+0.33)    \\

LP++-e        & - & - & 41.93 & 75.38 & 84.19 & 60.14 & - & 72.43  & 78.54  & 85.01 &  - & \multicolumn{1}{l|}{94.81} & 74.05    \\ 
LP++-h        & - & - & 42.58 & 75.91 & 84.72 & 59.67 & - & 71.96  & 75.81  & 85.63 &  - & \multicolumn{1}{l|}{95.32} &  73.95 (-0.10)   \\ 

ProText-e  & - & - & 47.72 & 76.42 & 82.67 & 64.18 & - & \underline{73.59}  & 75.68  & 86.28 &  - & \multicolumn{1}{l|}{94.89} & 75.18    \\ 
ProText-h  & - & - & 47.19 & 76.95 & 82.14 & 63.77 & - & 73.06  & 78.23  & 86.04 &  - & \multicolumn{1}{l|}{95.71} & 75.39 (+0.21)   \\ 
\cline{1-14}

IOTA-e & - & - & \underline{54.19} & \underline{77.31} & \underline{89.77} & \underline{74.83} & - & 73.55 & \underline{86.25} & 86.75 & - &  \multicolumn{1}{l|}{ \underline{96.32}}  & \underline{79.87} \\

IOTA-h & - & - & \textbf{58.45} & \textbf{79.37} & \textbf{91.01} & \textbf{77.08} & - & \textbf{74.47} & \textbf{87.70} & \textbf{88.40} & - &  \multicolumn{1}{l|}{ \textbf{97.19}}  & \textbf{81.70 (+1.83)} \\
\hline

CoOp-e          & 95.62 & 79.11 & 41.49 & 72.27 & 81.42 & 56.70 & 68.99 & 71.22 & 77.83 & 84.15 & 81.10 & \multicolumn{1}{l|}{93.14} &  75.25 &\multirow{14}{*}{\textbf{8/stage}} \\
CoOp-h         & 95.25 & 78.88 & 41.31 & 72.56 & 81.30 & 55.34 & 62.93 & 70.48 & 72.17 & 82.64 & 79.25 & \multicolumn{1}{l|}{89.05} &  73.43 (-1.82) \\
Co-CoOp-e       & 86.60 & 72.30 & 32.49 & 70.96 & 80.67 & 55.30 & 62.45 &    70.87   & 68.00 & 80.27 & 78.30 & \multicolumn{1}{l|}{87.71} &  70.49 \\
Co-CoOp-h      & 88.39 & 72.38 & 33.51 & 70.17 & 76.24 & 54.35 & 59.31 &   70.25    & 63.83 & 78.36 & 77.61 & \multicolumn{1}{l|}{88.08} &  69.37 (-1.12) \\
Clip-Adapter-e  & 95.74 & 76.53 & 42.33 & 69.51 & 80.52 & 54.28 & 69.41 &   69.37    & 76.99 & \underline{85.79} & 82.21 & \multicolumn{1}{l|}{94.03}  &  74.73 \\
Clip-Adapter-h & 95.41 & 76.63 & 43.26 & 70.75 & 81.30 & 55.36 & 69.79 &    68.93   & 74.93 & 85.51 & 82.63 & \multicolumn{1}{l|}{94.36}  &  74.91 (+0.18) \\
ProGrad-e       & 94.80 & 72.07 & 36.48 & 69.38 & 79.88 & 54.07 & 65.11 &  71.29& 75.11 & 82.41 & 74.20 & \multicolumn{1}{l|}{90.16} &    72.08  \\
ProGrad-h      & 94.32 & 76.92 & 41.67 & 72.03 & 81.09 & 56.36 & 66.60 &  70.03& 70.89 & 82.58 & 80.10 & \multicolumn{1}{l|}{90.62} &   73.60 (+1.52)   \\
Maple-e & 89.85 & 73.91 & 35.16 & 72.19 & 80.93 & 60.64 & 66.76 & 71.36 & 70.58 & 80.23 & 79.22 & \multicolumn{1}{l|}{87.58} &   72.37  \\ 
Maple-h & 91.47 & 75.75 & 36.12 & 72.94 & 80.37 & 60.98 & 68.03 & 70.72 & 75.37 & 81.02 & 79.75 & \multicolumn{1}{l|}{90.84} &  73.61 (+1.24)   \\ 
KgCoOp-e & 94.97 & 76.23 & 38.70 & 73.88 & 81.92 & 57.58 & 67.93 & 71.17 & 76.73 & 84.08 & 82.02 & \multicolumn{1}{l|}{91.89} & 74.76  \\ 
KgCoOp-h & 95.62 & 76.79 & 39.12 & 73.31 & 82.26 & 57.89 & 68.03 & 70.73  & 76.65 & 84.47 & 82.05 & \multicolumn{1}{l|}{92.19} & 74.93 (+0.17)   \\ 

TCP-e  & 94.68 & 76.02 & 39.27 & 73.26 & 82.67 & 61.89 & 68.84 &  71.46& 75.63 & 84.91 & 80.93 & \multicolumn{1}{l|}{92.34} &    75.16  \\
TCP-h  & 95.12 & 77.13 & 41.36 & 73.92 & 83.12 & 61.45 & 69.57 &  71.03& 74.99 & 85.36 & 81.58 & \multicolumn{1}{l|}{93.07} &   75.64 (+0.48)   \\

LP++-e & 94.77 & 76.48 & 39.18 & 73.15 & 82.94 & 60.18 & 67.39 & 71.96 & 74.27 & 81.09 & 80.62 & \multicolumn{1}{l|}{92.83} & 74.57  \\ 
LP++-h & 95.38 & 75.97 & 40.94 & 73.69 & 83.41 & 60.77 & 68.71 & 71.79  & 75.89 & 81.42 & 81.47 & \multicolumn{1}{l|}{92.12} & 75.13 (+0.56)   \\ 

ProText-e & 95.57 & 76.84 & 43.72 & \underline{74.11} & 83.76 & 61.27 & 68.95 & \underline{72.07} & 76.22 & 84.73 & 80.64 & \multicolumn{1}{l|}{94.41} & 76.02  \\ 
ProText-h & 96.03 & 76.92 & 44.88 & 73.58 & 83.47 & 62.84 & 69.41 & 71.55  & 76.84 & 84.28 & 81.36 & \multicolumn{1}{l|}{94.75} & 76.33 (+0.31)   \\ 
\cline{1-14}

IOTA-e & \underline{96.59} & \underline{82.33} & \underline{46.32} & 74.02 & \underline{88.12} & \underline{70.94} & \underline{70.17} & 71.61 & \underline{78.91} & 85.47 & \underline{83.24} &  \multicolumn{1}{l|}{ \underline{95.64}}  & \underline{78.61} \\

IOTA-h & \textbf{97.73} & \textbf{84.96} & \textbf{52.30} & \textbf{77.40} & \textbf{89.93} & \textbf{74.21} & \textbf{71.56} & \textbf{72.99} & \textbf{84.56} & \textbf{87.17} & \textbf{85.38} &  \multicolumn{1}{l|}{ \textbf{96.43}}  & \textbf{81.22 (+2.61)} \\

\hline
CoOp-e & 93.36 & 74.05 & \underline{37.29} & 69.94 & 80.89 & 54.17 & 64.21 &  69.72  & 72.32 & 80.57 & 79.06 & \multicolumn{1}{l|}{91.23} & \underline{72.23} &\multirow{14}{*}{\textbf{4/stage}}\\ 
CoOp-h & 92.53 & 75.50 & 36.99 & 71.15 & 72.13 & 52.81 & 58.56 &  69.81  & 61.63 & 75.62 & 76.71 & \multicolumn{1}{l|}{85.41} &  69.07 (-3.16) \\
Co-CoOp-e  & 82.99 & 70.49 & 32.19 & 69.66 & 79.70 & 54.11 & 59.73 &  69.32  & 58.38 & 77.03 & 75.20 & \multicolumn{1}{l|}{84.93} &  67.81 \\
Co-CoOp-h & 84.57 & 70.59 & 33.27 & 69.36 & 77.69 & 53.77 & 58.03 &  69.08  & 59.69 & 78.54 & 75.36 & \multicolumn{1}{l|}{83.60} &  67.80 (-0.01) \\ 
Clip-Adapter-e & 93.50 & \underline{74.27} & 36.84 & 66.15 & 80.33 & 53.10 & 64.13 & 68.54 & \underline{73.64} & 77.82 & \underline{79.71} & \multicolumn{1}{l|}{91.88}  &  71.66 \\
Clip-Adapter-h & 93.34 & 74.78 & 37.47 & 67.80 & 80.35 & 54.01 & 63.72 & 68.84 & 70.60 & 77.78 & 79.83 & \multicolumn{1}{l|}{92.17}  & 71.72 (+0.06)\\
ProGrad-e & 92.41 & 68.96 & 31.98 & 67.37 & 79.87 & 52.79 & 61.28 &    69.55   & 67.89 & \underline{80.77} & 72.43 & \multicolumn{1}{l|}{87.70} &   69.42   \\
ProGrad-h & 91.76 & 73.97 & 37.68 & 69.60 & 80.91 & 55.41 & 58.83 &   69.34    & 66.20 & 78.78 & 76.29 & \multicolumn{1}{l|}{85.66} &     70.37 (+0.95) \\
Maple-e & 82.42 & 70.81 & 33.45 & 70.76 & 80.07 & 57.74 & 57.23 & 69.53 & 52.09 & 75.63 & 76.26 & \multicolumn{1}{l|}{81.20} &  67.27   \\ 
Maple-h & 84.77 & 71.38 & 33.72 & 70.92 & 79.02 & 58.10 & 57.98 & 69.33 & 60.07 & 76.80 & 76.79 & \multicolumn{1}{l|}{80.43} &   68.28 (+1.01) \\ 
KgCoOp-e & 92.73 & 72.70 & 34.11 & 71.68 & 81.38 & 55.61 & 63.46 & 69.77 & 63.96 & 80.08 & 78.60 & \multicolumn{1}{l|}{88.72} &  71.90   \\ 
KgCoOp-h & 93.18 &72.83  & 35.29 & 71.59 & 82.10 & 56.16 & 63.35 & 69.92 & 71.53 & 80.29 & 78.76 & \multicolumn{1}{l|}{89.33} &   72.03 (+0.13) \\ 

TCP-e  & 90.85 & 71.89 & 34.97 & 69.74 & 80.11 & 56.76 & 61.14 &  69.11& 65.94 & 78.73 & 77.34 & \multicolumn{1}{l|}{89.42} &    70.50  \\
TCP-h  & 91.33 & 73.24 & 37.90 & 71.45 & 81.73 & 58.13 & 62.75 &  69.84& 69.81 & 79.19 & 77.61 & \multicolumn{1}{l|}{90.81} &   71.98 (+1.48)   \\

LP++-e & 91.12 & 71.53 & 34.88 & 69.04 & 80.98 & 56.95 & 61.89 & 69.26 & 66.07 & 80.24 & 76.82 & \multicolumn{1}{l|}{88.76} &   70.63  \\ 
LP++-h & 92.47 & 72.58 & 36.15 & 70.12 & 82.45 & 57.81 & 62.21 & 69.59 & 68.62 & 79.88 & 77.45 & \multicolumn{1}{l|}{90.13} &  71.62 (+0.99)   \\ 

ProText-e & 90.29 & 72.91 & 35.49 & \underline{71.99} & 80.33 & 57.11 & 63.88 & \underline{69.86} & 70.18 & 78.92 & 77.92 & \multicolumn{1}{l|}{90.95} & 71.65  \\ 
ProText-h & 91.08 & 74.31 & 39.72 & 70.47 & 81.86 & 58.42 & 62.67 & 69.98  & 67.43 & 79.83 & 78.48 & \multicolumn{1}{l|}{91.38} & 72.14 (+0.49)   \\ 
\cline{1-14}

IOTA-e & \underline{93.91} & 72.43 & 35.28 & 67.11 & \underline{81.46} & \underline{64.23} & \underline{64.31} & 68.25 & 66.32 & 79.17 & 78.67 &  \multicolumn{1}{l|}{ \underline{93.09}}  & 72.02 \\

IOTA-h & \textbf{95.94} & \textbf{76.86} & \textbf{40.53} & \textbf{71.92} & \textbf{85.54} & \textbf{67.56} & \textbf{65.48} & \textbf{70.03} & \textbf{75.40} & \textbf{81.08} & \textbf{80.41} &  \multicolumn{1}{l|}{ \textbf{94.26}}  & \textbf{75.42 (+3.40)} \\
\bottomrule[1.5pt]
\end{tabular}}
\label{tbl:e5}
\end{table*}

\section{Experiments}
\begin{figure*}[t]
\centering
  	\includegraphics[width=1\linewidth]{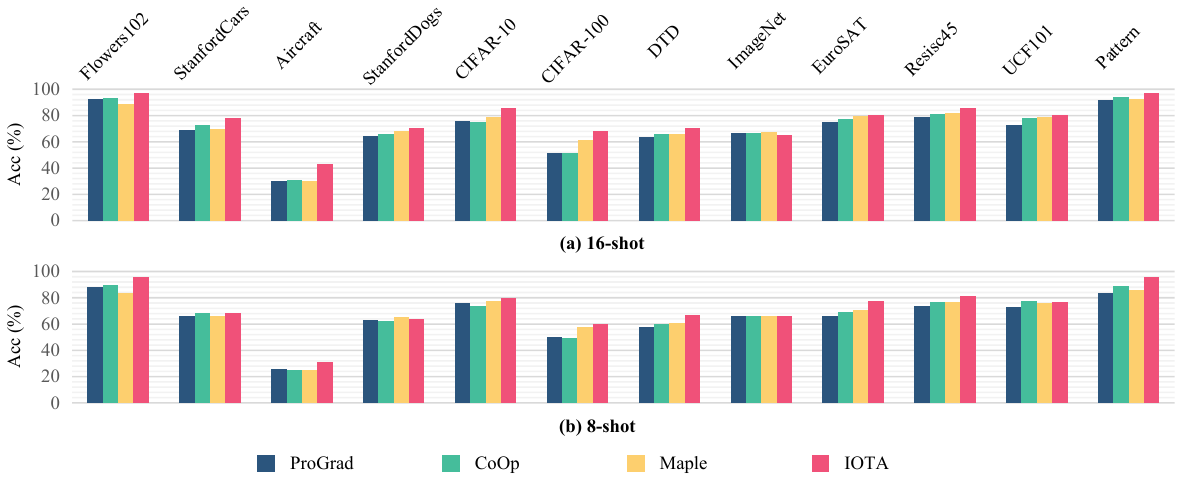}
  \caption
    { Comparison with the SOTA methods on 12 datasets under the \textbf{few-shot downstream task adaptation setting}, where the base learner is \textbf{ViT-B/32}. (a) 16-shot. (b) 8-shot.}
  \label{fig:vitb32}
\end{figure*}

\begin{figure*}[t]
\centering
  	\includegraphics[width=1\linewidth]{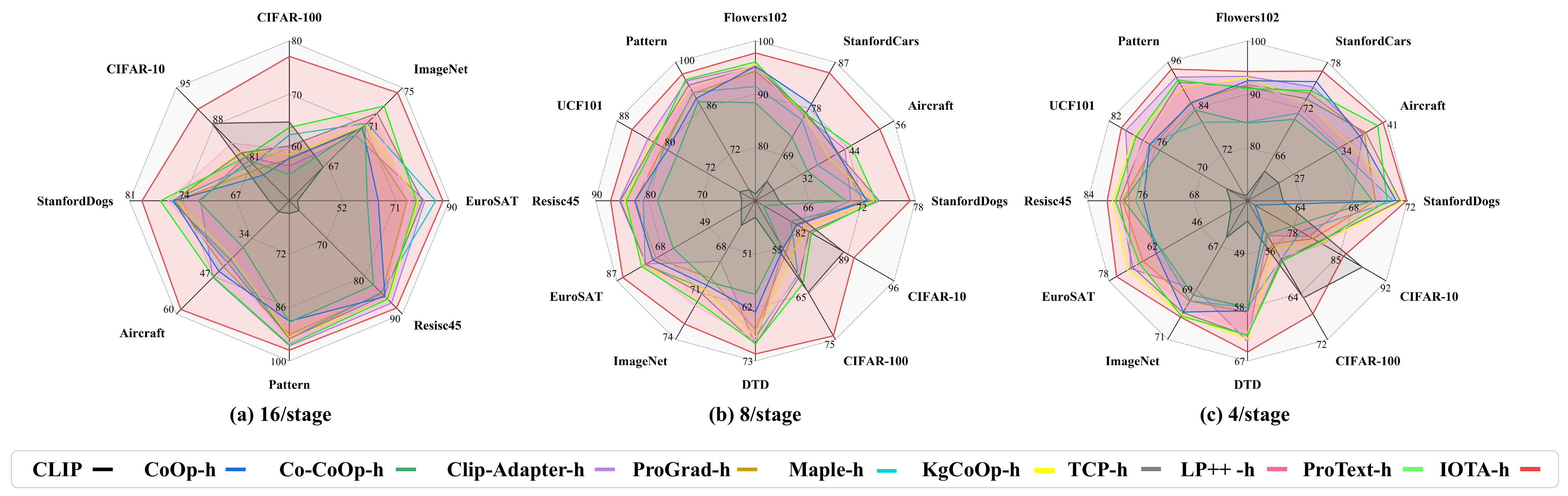}
  \caption
    { Performance comparison under the \textbf{easy-to-hard downstream task adaptation} setting, where the base learner is \textbf{ViT-B/16}. Our method performs best on almost all datasets under three easy-to-hard downstream task adaptation settings, namely (a) 16/stage, (b) 8/stage, and (c) 4/stage. }
  \label{fig:radar}
\end{figure*}

We conducted comprehensive experimental evaluations on
12 benchmark datasets under two settings: (1) Few-shot downstream task adaptation; (2) Easy-to-hard downstream task adaptation. 
In the following part of this section, we first
described the 12 benchmark datasets, provided the experimental setup of the proposed method, and then presented the evaluation results and their analyses, finally showed ablation studies and visualization results. 

\subsection{Experimental Setup}

\subsubsection{Datasets}
For few-shot downstream task adaptation and easy-to-hard downstream task adaptation, we conducted experimental evaluations on three kinds of (1) \textbf{Natural datasets}: CIFAR-10~\cite{cifar}, CIFAR-100~\cite{cifar}, DTD~\cite{dtd}, and ImageNet~\cite{imagenet}. (2) \textbf{Fine-grained datasets}: Flowers102~\cite{flower102},  StanfordCars~\cite{cars}, FGVCAircraft~\cite{aircraft}, and StanfordDogs~\cite{dogs}. (3) \textbf{Specialized datasets}: EuroSAT~\cite{sat}, Resisc45~\cite{resisc}, UCF101~\cite{ucf}, and Pattern~\cite{pattern}. 
Detailed statistics for these datasets are listed in Table~\ref{tbl:a1}. 
For StanfordDogs, CIFAR-10, CIFAR-100, DTD, ImageNet, Resisc45, and Pattern, we followed the official dataset split strategy. 
For Flowers102, StanfordCars, Aircraft, EuroSAT, and UCF, 
we followed the split strategy used in Coop. 
If a dataset contains a validation set, we integrate the validation set into the train set.

\subsubsection{Implementation Details}

All experiments were implemented in PyTorch and conducted on two NVIDIA RTX 3090 GPUs. 
Each result is reported as the average over three independent runs. 

\textbf{Model Architecture.} 
We adopt the pre-trained CLIP visual encoders ViT-B/16 and ViT-B/32 as base learners. 
Both variants share the same transformer architecture, consisting of 12 layers, a hidden dimension of 768, and 12 attention heads. 
Compared with ViT-B/32, ViT-B/16 provides finer-grained image representations and generally achieves better downstream performance. 
The CLIP semantic encoder is used as the knowledge encoder. 

\textbf{Model Scale and Efficiency.} 
Our models contain an average of 153.4M parameters, of which 149.6M belong to the frozen base learner and knowledge encoder, and only 3.8M (approximately 2.5\%) are trainable. 
The computational complexity is 129.9 GFLOPs, measured at an input resolution of 224×224. 

\textbf{Training Configuration.} 
All models are optimized with SGD using a mini-batch size of 32 or 64. 
The initial learning rate is set to 0.003 and gradually decayed to 0.0001 via a cosine annealing schedule for the few-shot task adaptation and the easy stage of the easy-to-hard curriculum. 
During the hard curriculum stage, the learning rate starts from 0.0005 and is annealed to 0.0001 in the same manner. 
Each model is trained for 50 epochs, following the same data augmentation strategies as our comparison baselines. 

\textbf{Hyper-parameters.} 
The hyper-parameters $\lambda$ and $\tau$ are set to 0.2 and 0.1, respectively. 
The length of the learnable prompt $\mathbf{v}$ is set to 2. 

\subsection{Comparison Results}
\subsubsection{Comparison Methods}
We compared our model with the SOTA prompt learning methods (\ie CoOp~\cite{coop}, Co-CoOp~\cite{cocoop}, Maple~\cite{maple}, KgCoOp~\cite{kgcoop}, ProGrad~\cite{prograd}, TCP~\cite{TCP}, LP++~\cite{lp24}, and ProText~\cite{ProText}). 
In addition,
we also compared with the representative adapter-based method (\ie Clip-Adapter~\cite{clip-adapter}) and the Zero-shot CLIP~\cite{clip}. 
For a fair comparison, 
we adopted the same pre-trained CLIP as the base model of our comparison methods, and we used the dataset split strategy as the same as our method to reproduce these methods.

\subsubsection{Experimental Setting}

\textbf{Few-shot Downstream Task Adaptation.} 
We evaluate model adaptability in few-shot scenarios by fine-tuning models with a small number of randomly selected training samples and testing on the full evaluation set. 
For fair comparison, all methods are trained with the same sampled instances.

\textbf{Easy-to-Hard Downstream Task Adaptation.} 
To further assess the effectiveness of corrective knowledge, we adopt a two-stage curriculum learning strategy, where each class provides 16, 8, and 4 samples per stage, respectively. 
Due to the limited number of available training samples, the 16-sample-per-stage setting is excluded from Flowers102, StanfordCars, DTD, and UCF101, as these datasets do not contain enough qualified “hard” samples after the easy stage.

\subsubsection{Results of Few-Shot Downstream Task Adaptation}
For the few-shot downstream task adaptation, 
we compared our model with all the SOTA methods on 12 benchmark datasets with 16-shot and 8-shot settings. 

As shown in Table~\ref{tbl:e1} and Fig.~\ref{fig:concept} (b), our IOTA outperforms the SOTA methods with the 16-shot setting: (1) it has achieved the highest accuracies on all benchmark datasets. (2) it has achieved the highest average accuracy.  Compared to the best prompt-based method ProText, our method surpasses it by 4.70\% in average accuracy. Also, our method surpasses the adapter-based method Clip-Adapter by 6.09\% in average accuracy. 
Besides, our IOTA also outperforms the SOTA methods with the 8-shot setting: (1) It has achieved the highest accuracies or compatible ones on all benchmark datasets; (2) it has achieved the highest average accuracy.  
Compared to the best method ProText, our IOTA surpasses it by 2.13\% in average accuracy. 

Moreover, as we can see from Table~\ref{tbl:a4}, our method also has achieved the highest average accuracy when we use pre-trained ViT-B/32 as the base learner: (1) IOTA surpasses the best methods by 4.68\%, 3.87\%, and 2.02\% on 4 fine-grained datasets, 4 natural datasets, and 4 specialized datasets with the 16-shot setting, respectively; 
(2) IOTA outperforms the best methods with a gap of 2.09\%, 1.70\%, and 3.12\% on 4 fine-grained datasets, 4 natural datasets, and 4 specialized datasets with the 8-shot setting, respectively. 
The more detailed results of each dataset are shown in Fig.~\ref{fig:vitb32}. 

\subsubsection{Results of Easy-to-Hard Downstream Task Adaptation}
According to Table~\ref{tbl:e5} and Fig.~\ref{fig:radar}, our method demonstrates clear advantages over the SOTA methods: (1) Our IOTA-h has obtained the highest accuracies across all datasets and the highest average accuracies for the 16/stage, 8/stage, and 4/stage settings; (2) our IOTA-e has achieved the highest average accuracies for the 16/stage, 8/stage, and 4/stage settings; (3) our IOTA-e has achieved the highest accuracies or compatible ones on all datasets across the three settings; (4) our IOTA-h outperforms our IOTA-e by 1.83\%, 2.61\%, and 3.40\%  for the three settings, respectively, which is the most significant improvement compared to the SOTA methods. 
In conclusion, IOTA-h shows substantial improvement over IOTA-e, validating the effectiveness of corrective knowledge.  
Besides, we observe that the advantage of IOTA-e becomes less evident when the number of training samples is very limited, likely due to the insufficient corrective knowledge available under such conditions. 

\subsection{Ablation Studies and Visualization}
\begin{table}[t]
\caption
		{
       Ablation studies of our White Box module on 12 datasets. B-B, W-B, and VPT indicate the Black Box module, the White Box module, and the visual prompt tuning, respectively. The base learner is \textbf{ViT-B/16}. 
		}
	\centering
\begin{tabular}{l|lll|l}
\toprule[1.5pt]
Method  & \multicolumn{1}{l|}{B-B} & \multicolumn{1}{l|}{W-B} & VPT & \multicolumn{1}{l}{Avg. Acc.} \\ \hline
B-B + linear probe & \checkmark & \ding{55}  & \ding{55}  &  76.53   \\
\hline
Our w/o W-B & \checkmark & \ding{55}  & \checkmark  &  77.38   \\
\hline
Our method &  \checkmark  &  \checkmark & \checkmark  & \textbf{81.71} \\ \bottomrule[1.5pt]
\end{tabular}

	\label{tbl:ab}
\end{table}

\begin{table}[t]
\caption
		{
        Comparison of the average accuracy with different settings of hyper-parameter $\lambda$ on all 12 datasets. The visual encoder is \textbf{ViT-B/16}. 
		}
\small
\centering
\begin{tabular}{c|ccccc}
\toprule[1.5pt]
$\lambda$  & 0.1 & 0.2 & 0.5 & 1.0 & 2.0 \\ \hline
$Avg. acc. \uparrow$   &  81.48  & \textbf{81.71}   & 81.39 & 81.35 & 81.11  \\ \bottomrule[1.5pt]
\end{tabular}
		\label{tbl:e4}
\end{table}

\begin{table}[htbp]
\caption
		{
       Evaluating our method on the wrong-predicted samples of our base learner (the accuracy of our base learner on these samples is 0.00). This experiment is conducted on all 12  datasets under the \textbf{16-shot downstream task adaptation setting}.  The base learner is \textbf{ViT-B/16}. 
		}
	\centering
\begin{tabular}{l|c}
\toprule[1.5pt]
Method   & \multicolumn{1}{l}{Average Accuracy} \\ \hline
Our base learner &  0.00 \\
\hline

Co-CoOp &  49.27 \\

Clip-Adpter &  54.02 \\

Maple &  55.47 \\

KgCoOp &  56.39 \\

CoOp &  57.48 \\

ProText &  58.36 \\

TCP &  58.73 \\
\hline
IOTA  & \textbf{65.81} \\ \bottomrule[1.5pt]
\end{tabular}
\label{tbl:ab3}
\end{table}

\begin{figure}[t]
\centering
  	\includegraphics[width=1.0\linewidth]{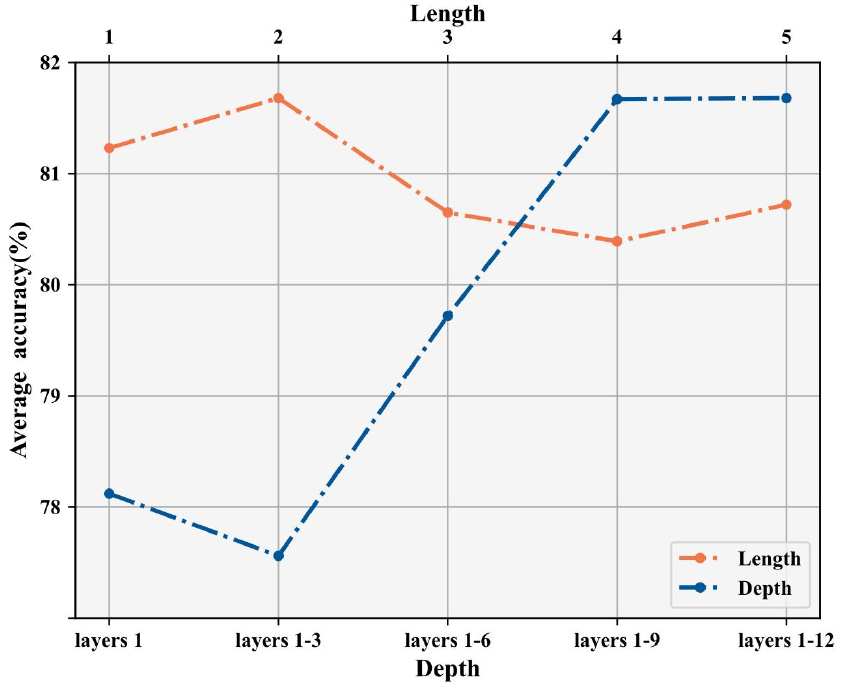}
  \caption
    {
	 Ablation studies of the length and depth of our learnable prompt $\textbf{v}$ on all 12 datasets, where the base learner is \textbf{ViT-B/16}. }
  \label{fig:ab}
\end{figure}

\begin{figure}[t]
\centering
\includegraphics[width=0.9\linewidth]{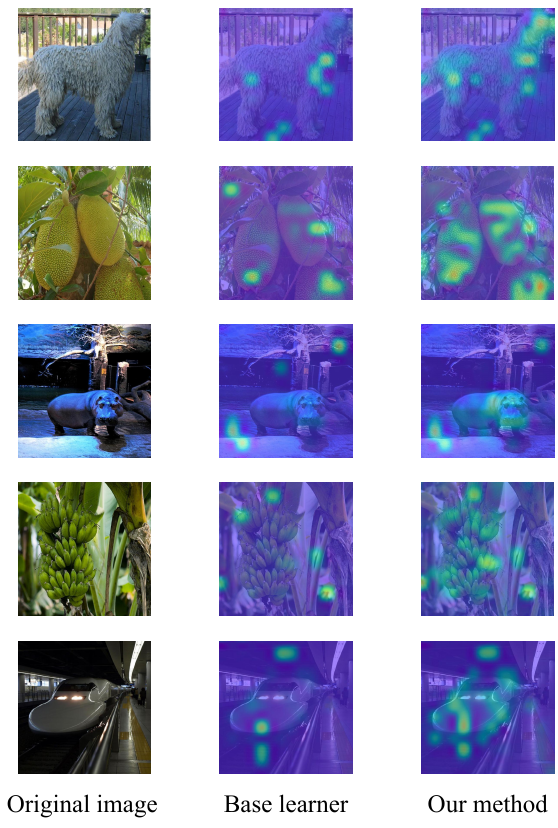}
  \caption
    {Self-attention map visualization of the comparison between our base learner and our method on \textbf{ImageNet} under the \textbf{few-shot downstream task adaptation setting}. We look at the self-attention of the CLS token on the heads of the last layer. } 
  \label{fig:att}
\end{figure}

\begin{figure}[t]
\centering
  	\includegraphics[width=0.8\linewidth]{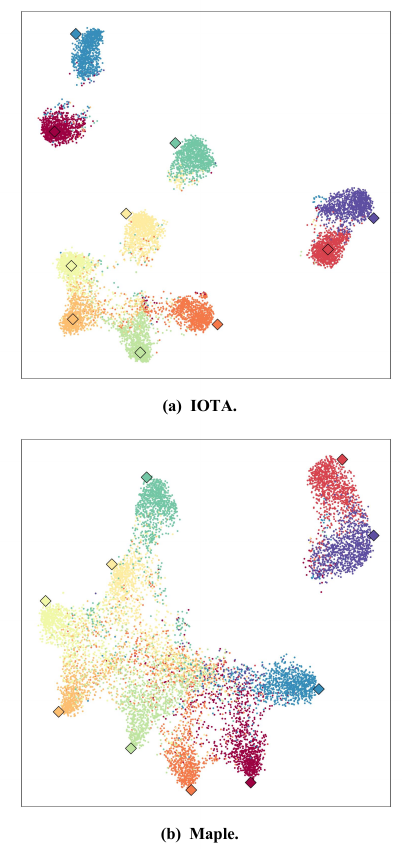}
  \caption
    {
	 T-SNE visualization of the comparison between IOTA and Maple on \textbf{Cifar-10} under the \textbf{few-shot downstream task adaptation setting}.  In our method, the MATCH tokens of 10 classes are well separated and distributed around the corresponding human prompts (represented by diamonds).}
  \label{fig:tsne}
\end{figure}

To further investigate the inner workings of our method, we designed ablation studies on the White Box module, hyper-parameter $\lambda$, prompt length, prompt depth, and ability to correct wrong predictions of our base learner.  
All experiments in ablation studies are conducted on the 16-shot downstream task adaptation setting. 
In addition,
we reported visualization results to provide intuitive explanations for our method.  
\subsubsection{Effectiveness of the White Box Module}
To dive into the effectiveness of our White Box module, we removed the White Box module and adopted visual prompt tuning~\cite{vpt} and linear probe~\cite{clip} to fine-tune the Black Box module, respectively, which are commonly used in downstream task adaptation. 
As shown in TABLE~\ref{tbl:ab}, our method achieves the best performance compared with ‘Our w/o W-B’ and ‘B-B + linear probe’, leading to improvements of 4.33\% and 5.18\% in average accuracy, respectively. 
These results demonstrate that the White Box module is helpful for downstream task adaptation under the guidance of corrective knowledge. 

\subsubsection{Impact of the Hyper-parameter $\lambda$} 
The hyper-parameter $\lambda$ controls the contribution of the corrective knowledge-guided constraint $\mathcal{L}_{ckg}$. 
We changed the value $\lambda$ from 0.1 to 2. 
As illustrated in Table~\ref{tbl:e4}, our method achieves the best performance when $\lambda = 0.2$. 

\subsubsection{Impact of the Prompt Length}
Fig.~\ref{fig:ab} shows how the length of the learnable prompt $\textbf{v}$ affects our method.  
As the prompt length increases, the model performance reaches its peak at length = 2. Beyond this point, further increasing the length results in a slight decline in performance. The observation is reasonable and aligns with prior findings~\cite{vpt}, which suggest that longer prompts do not always yield better results. Excessive prompt length increases model complexity and overfitting risk, both of which can hinder performance. 

\subsubsection{Impact of the Prompt Depth}
Fig.~\ref{fig:ab} illustrates the effect of the depth of the learnable prompt $\textbf{v}$ for our method. 
For example, 
‘depth: layer 1-6’ means that we inserted $\textbf{v}$ into the 1st - 6th blocks of our Black Box module according to Eq. (10). 
We observed that (1) our method achieves the best performance when the prompt depth is set to 1-12 and (2) in most cases, deeper prompts consistently yield better results compared to shallower ones. This observation aligns with the general consensus that deeper prompts lead to better performance. The slight performance degradation observed when moving from layer 1 to layers 1-3 is likely due to the limited number of trainable parameters in both configurations, which increases sensitivity to optimization variance and results in small, yet acceptable, fluctuations in performance. Notably, with further increases in prompt depth, the model’s performance improves steadily and consistently.

\subsubsection{Ability to Correct Wrong Predictions of Our Base Learner} 
In this part, 
we designed an ablation study to evaluate the ability of our method to correct wrong predictions of our base learner. 
We evaluated our method on the wrong-predicted samples of our base learner, where the average accuracy of our base learner on these samples is 0.00. 
As we can see from Table~\ref{tbl:ab3}, 
our method outperforms ‘Our base learner’ with a gap of 65.81\%  on the average accuracy, 
which means 65.81\% of wrong predictions are corrected by our method under the 16-shot downstream task adaptation setting. 
It proves our method can use few-shot samples to correct most of the wrong predictions of our base learner. 
Moreover, our method exhibits superior performance compared to other baslines, thereby highlighting the effectiveness of corrective knowledge. 

\subsubsection{Visualization}
First, 
we visualized the self-attention maps of our base learner and our method on ImageNet. 
The self-attention map indicates the self-attention of the CLS token and other image tokens on the heads of the last layer. 
Note that our IOTA uses the NCLS token to replace the CLS token. 
As shown in Fig.~\ref{fig:att}, the NCLS token of our IOTA pays more attention to the object region, indicating that our IOTA is better adapted to the downstream task. 

Besides, 
to verify whether the $\mathcal{L}_{ckg}$ and MATCH token can improve the matching accuracy of prompt selection, we visualized all MATCH tokens and human prompts of our method. 
Meanwhile, we visualized all vision prompt tokens and language prompt tokens of Maple, which are used to match each other. Comparing Fig.~\ref{fig:tsne} (a) and Fig.~\ref{fig:tsne} (b), we observed that matching results between our MATCH tokens and human prompts are significantly better.

\section{Conclusions And Future Works}
In this work, 
we propose a Black-White Box prompt learning framework for downstream task adaptation. 
The framework integrates a pre-trained visual encoder as the Black Box module and a knowledge-guided prompt learning mechanism as the White Box module.
Within the White Box module, corrective knowledge is explicitly defined and verbalized through true-wrong prompt pairs.
A corrective knowledge-guided prompt selection strategy is further introduced to retrieve suitable prompts from the knowledge prompt set, thereby refining the base learner’s representations. 
We demonstrate the effectiveness of corrective knowledge and the superiority of our method across 12 benchmark datasets. 
An intriguing direction for future research involves extracting rules from external knowledge graphs to develop more interpretable White Box modules. 
\bibliographystyle{ieee_fullname}


\end{document}